\documentclass{article}
\usepackage{ijcai26}

\usepackage{times}

\usepackage[utf8]{inputenc}
\usepackage{url}
\usepackage[hidelinks]{hyperref}
\usepackage{amsmath}
\usepackage{amssymb}
\usepackage{amsthm}

\usepackage[small]{caption}
\usepackage{graphicx}

\usepackage{algorithm}
\usepackage{algorithmic}

\usepackage[switch]{lineno}

\usepackage{booktabs}
\usepackage{multirow}
\usepackage{makecell}
\usepackage{colortbl}
\usepackage{arydshln}
\usepackage{soul}
\usepackage{siunitx}
\usepackage{bm}
\usepackage{float}
\usepackage[table]{xcolor}
\definecolor{sciBlue}{RGB}{0, 114, 178}  
\definecolor{sciRed}{RGB}{213, 94, 0}    
\definecolor{sciGreen}{RGB}{0, 158, 115} 
\definecolor{sciGray}{RGB}{100, 100, 100} 
\definecolor{softGrid}{gray}{0.9}        
\definecolor{shadegray}{gray}{0.95}      
\definecolor{problue}{RGB}{0, 90, 180}
\definecolor{prored}{RGB}{200, 50, 50}
\definecolor{rowgray}{gray}{0.95}
\definecolor{cvprblue}{rgb}{0.93, 0.95, 1.0}
\definecolor{highlightrow}{RGB}{235, 242, 250}
\definecolor{graytext}{RGB}{120, 120, 120}
\definecolor{deepblue}{HTML}{005EB8}
\definecolor{brickred}{HTML}{C23B22}
\definecolor{plotbg}{HTML}{FAFAFA}
\definecolor{highlightgray}{gray}{0.92}
\definecolor{rulecolor}{gray}{0.9}
\definecolor{slategray}{rgb}{0.44, 0.5, 0.56}
\definecolor{deepgreen}{HTML}{1B9E77}
\definecolor{deeporange}{HTML}{E69F00} 

\usepackage{tikz}
\usetikzlibrary{shadows,positioning,calc,backgrounds,shapes,arrows.meta}
\definecolor{gaincolor}{RGB}{0,140,0}
\definecolor{losscolor}{RGB}{194,59,34}
\definecolor{bbgrayblue}{RGB}{245,248,252}
\newcommand{\up}[1]{%
  \textcolor{gaincolor}{\scalebox{0.99}{\scriptsize$\uparrow$\,#1}}%
}

\usepackage{placeins}

\usepackage{pgfplots}
\pgfplotsset{compat=1.18}
\pgfplotsset{
  colormap/viridis
}
\usepgfplotslibrary{groupplots}

\usepackage{pifont}

\title{DiEC: Diffusion Embedded Clustering}
\author{
Haidong Hu$^{2}$
\and
Xiaoyu Zheng$^{2}$
\and
Jin Zhou$^{1,2,}$\footnote{Corresponding author: Jin Zhou}
\and
Yingxu Wang$^{2}$
\and
Rui Wang$^{1}$
\and
Pei Dong$^{2}$
\and
\\Shiyuan Han$^{1}$
\and
Lin Wang$^{2}$
\and
C.~L.~Philip Chen$^{3}$
\and
Tong Zhang$^{3}$
\And
Yuehui Chen$^{2}$\\
\affiliations
$^1$School of Artificial Intelligence, Shandong Women’s University, China\\
$^2$Shandong Key Laboratory of Ubiquitous Intelligent Computing, University of Jinan, China\\
$^3$School of Computer Science and Engineering, South China University of Technology, China\\
\emails
haidonghu.ml@gmail.com, zhoujin@sdwu.edu.cn
}

\begin{document}

\maketitle

\begin{abstract}
Deep clustering methods typically rely on a single, well-defined representation for clustering. In contrast, pretrained diffusion models provide abundant and diverse multi-scale representations across network layers and noise timesteps. However, a key challenge is how to efficiently identify the most clustering-friendly representation in the layer$\times$timestep space. To address this issue, we propose Diffusion Embedded Clustering (DiEC), an unsupervised framework that performs clustering by leveraging optimal intermediate representations from pretrained diffusion models. DiEC systematically evaluates the clusterability of representations along the trajectory of network depth and noise timesteps. Meanwhile, an unsupervised search strategy is designed for recognizing the Clustering-optimal Layer (COL) and Clustering-optimal Timestep (COT) in the layer$\times$timestep space of pretrained diffusion models, aiming to promote clustering performance and reduce computational overhead. DiEC is fine-tuned primarily with a structure-preserving DEC-style KL-divergence objective at the fixed COL + COT, together with a random-timestep diffusion denoising objective to maintain the generative capability of the pretrained model. Without relying on augmentation-based consistency constraints or contrastive learning, DiEC achieves excellent clustering performance across multiple benchmark datasets.
\end{abstract}

\begin{figure}[t]
  \centering
  \includegraphics[
    page=1,
    width=\columnwidth,
    trim=2mm 2mm 2mm 2mm, 
    clip
  ]{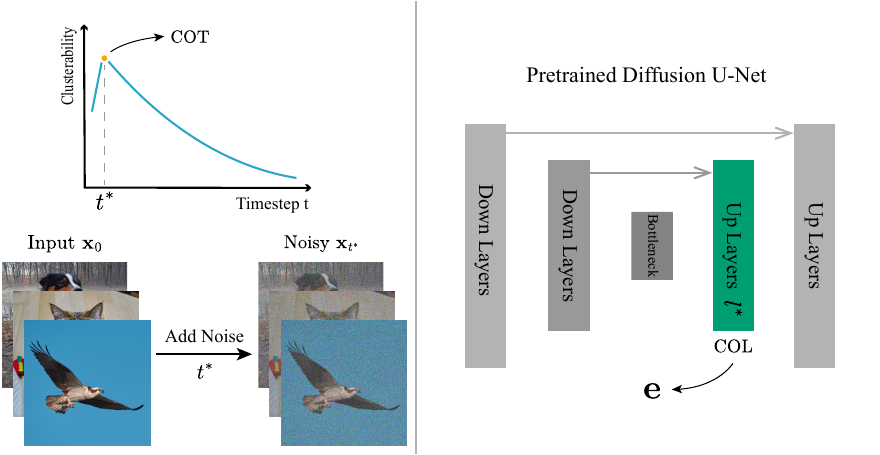}
\caption{
    \textbf{Schematic of the feature extraction process in DiEC.}
    A clean image \(x_0\) is perturbed with noise at the selected Clustering-Optimal Timestep (COT, \(t^*\)) to produce a noisy input \(x_{t^*}\). This input, conditioned on the timestep embedding, is fed into the pretrained diffusion U-Net. The activation from the dynamically selected Clustering-Optimal Layer (COL, \(l^*\)) is subsequently extracted as the embedding representation \(e\) for downstream clustering.
}
  \label{fig:teaser}
  \vspace{-2mm}
\end{figure}

\section{Introduction}
Clustering aims to uncover the intrinsic structure of unlabeled data. However, modern datasets are often high-dimensional, strongly nonlinear, and heavily contaminated by noise, making classical clustering methods that rely on shallow features and fixed distance metrics unreliable in practice \cite{liu2016multiple,liu2017sparse}. By coupling neural representation learning with clustering objectives, deep clustering learns more semantic and noise-robust embeddings, demonstrating good performance in applications such as image understanding, anomaly detection, and large-scale retrieval \cite{ren2024deep}.

Based on diverse network architectures, such as autoencoders (AE) \cite{xie2016unsupervised,guo2017improved}, variational autoencoders (VAE) \cite{jiang2016variational,yang2019deep}, and generative adversarial networks (GAN) \cite{mukherjee2019clustergan}, embedding-based deep clustering methods typically learn representations by jointly optimizing reconstruction and clustering objectives. Nevertheless, these methods ultimately rely on a single, fixed latent representation for clustering, which lacks the dynamic adaptability and struggles to align with structural variations of data across different semantic scales.

Diffusion models, a significant breakthrough in generative artificial intelligence, excel at precisely modeling complex data distributions and capturing multi-scale semantic representations. Through forward noising and reverse denoising, they construct hierarchical features across different timesteps. A recent study \cite{wang2024diffusion} further suggests that denoising representations can be spontaneously organized into multiple low-dimensional subspaces, naturally aligning with the needs of clustering. Yet, existing diffusion-based clustering research \cite{yan2025clusterddpm,uziel2025clustering,yang2024dific} has yet to fully exploit this potential, and in particular lack systematic evaluation and effective selection of clustering-friendly representations across network layers and timesteps. Moreover, the high computational and time cost of diffusion inference remains a critical practical constraint.

Motivated by these insights, we propose \textbf{Diffusion Embedded Clustering (DiEC)}, an unsupervised framework that leverages internal representations of a pretrained diffusion U-Net for clustering. DiEC views the layer$\times$timestep space as a representation trajectory and performs unsupervised evaluation to identify clustering-optimal readouts, namely the \textbf{Clustering-Optimal Layer (COL)} and the \textbf{Clustering-Optimal Timestep (COT)}. To alleviate interference between denoising and clustering objectives, DiEC introduces \textbf{residual feature decoupling} for effective task adaptation. For optimization, DiEC couples a DEC-style $\mathrm{KL}$ self-training objective with graph regularization to strengthen structural consistency, while retaining a random-timestep denoising reconstruction objective to preserve generative capability. Extensive experiments and ablations studies demonstrate that DiEC achieves excellent clustering performance on multiple benchmarks without relying on augmentation-based consistency constraints or contrastive learning. Our contributions are:
\begin{itemize}
\item DiEC provides a novel insight into diffusion-based clustering, revealing that effective feature selection within the multi-scale semantic representations of diffusion models is crucial for improving clustering performance.
\item DiEC further develops an efficient, cost-aware, and label-free search strategy for recognizing clustering-friendly COL + COT in the layer$\times$timestep space of pretrained diffusion models, aiming to promote clustering performance and reduce computational overhead.
\item DiEC is fine-tuned primarily with a structure-preserving DEC-style KL-divergence clustering objective at the fixed COL + COT, together with a random-timestep diffusion denoising objective to maintain the generative capability of pretrained diffusion models.
\end{itemize}

\section{Related Work}
\paragraph{Embedding-based Deep Clustering.}
Embedding-based deep clustering methods aim to uncover latent cluster structure by learning a low-dimensional representation jointly with clustering objectives, enabling end-to-end optimization. 
AE-based approaches, represented by DEC \cite{xie2016unsupervised} and its extensions \cite{guo2017improved,yang2019deep,cai2022efficient,zhang2024unsupervised,guo2017deep}, follow the classic ``encode--cluster--refine'' paradigm, typically combining reconstruction with clustering losses to preserve data structure while improving separability. 
However, purely deterministic encoders may overlook data uncertainty, and naively coupling reconstruction and clustering objectives can lead to optimization conflicts. 
VAE-based methods, such as VaDE \cite{jiang2016variational}, cast clustering as latent-variable inference by modeling embeddings as distributions and introducing mixture-based priors (e.g., GMM) in the latent space \cite{yang2019deepcluster}. 
Recent variants further extend this line to more challenging settings, including incomplete multi-view clustering and multimodal generative clustering \cite{xu2024deep,palumbo2024deep}. 
GAN-based approaches such as ClusterGAN \cite{mukherjee2019clustergan} leverage adversarial training and hybrid discrete--continuous latents to learn discriminative representations for clustering, while also inheriting practical issues of GAN training instability and the lack of explicit probabilistic inference. 
Overall, many existing pipelines rely on a fixed, single-scale readout for clustering, which limits flexibility in capturing structural variations across semantic scales. 
In contrast, our approach leverages diffusion models to construct a richer multi-scale representation space across network layers and diffusion timesteps, enabling more effective selection of clustering-friendly readouts.

\paragraph{Diffusion-Based Deep Clustering.}
Recent work \cite{wang2024diffusion} suggests that denoising representations in diffusion models can spontaneously organize data into multiple low-dimensional subspaces, indicating strong potential for clustering. Building on this insight, diffusion-based clustering studies have rapidly emerged, broadly falling into three directions. 
First, several methods inject clustering signals as conditional information into the diffusion generative process to promote class-consistent generation and more separable cluster structures \cite{yan2025clusterddpm,palumbo2024deep,yang2024dific,yang2022learning}. 
Second, other approaches treat the diffusion process as a feature extractor or a data augmentation mechanism to enhance clustering performance \cite{uziel2025clustering,qiu2024multi,zhu2025diffusion}. 
Third, diffusion-driven clustering has also been explored in practical application scenarios, such as hyperspectral imaging and ultrasound enhancement \cite{chen2023diffusion,chen2025diffusionclusnet}. 
In contrast to prior diffusion clustering methods, we focus on systematically identifying clustering-friendly readouts from the intrinsic multi-scale representations of pretrained diffusion models, aiming to improve clustering quality while reducing computational overhead.

\section{Background}

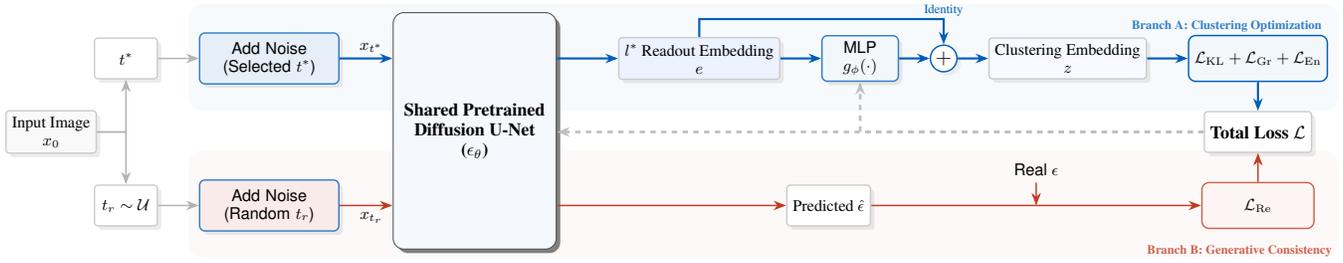
\begin{figure*}[t]
\centering
\resizebox{1\textwidth}{!}{%
\begin{tikzpicture}[
    node distance=1.2cm,
    >={Stealth[length=2.5mm]}, thick,
    font=\sffamily\small,
    tensor/.style={
        rectangle, rounded corners=2pt, draw=gray!40, fill=white,
        drop shadow={opacity=0.15}, minimum height=0.8cm, minimum width=1.2cm,
        align=center, font=\footnotesize
    },
    block/.style={
        rectangle, rounded corners=3pt, draw=deepblue!80, fill=deepblue!5,
        minimum height=1cm, minimum width=2cm, align=center,
        drop shadow={opacity=0.15}
    },
    unet/.style={
            rectangle, rounded corners=6pt,
            draw=black!70, fill=gray!10,
            minimum height=4.5cm, minimum width=3.2cm,
            align=center, font=\bfseries
        },
    loss/.style={
        rectangle, rounded corners=4pt, draw=brickred!80, fill=brickred!5,
        minimum height=0.9cm, minimum width=2.2cm, align=center,
        font=\footnotesize\bfseries
    },
    sum/.style={
        circle, draw=deepblue, fill=white, thick, inner sep=1pt,
        drop shadow={opacity=0.1}
    },
    blueflow/.style={->, draw=deepblue, very thick},
    redflow/.style={->, draw=brickred, thick},
    grayflow/.style={->, draw=gray!60, thick}
]

\node[tensor, fill=gray!5] (input) {Input Image\\$x_0$};

\node[block, above right=0.5cm and 2cm of input, fill=deepblue!10, text width=2.5cm] (noise_fix) {Add Noise\\(Selected $t^*$)};
\node[tensor, left=0.8cm of noise_fix, text width=1cm] (cond_fix) {$t^*$};

\node[block, below right=0.5cm and 2cm of input, fill=brickred!10, text width=2.5cm] (noise_rnd) {Add Noise\\(Random $t_r$)};
\node[tensor, left=0.8cm of noise_rnd, text width=1cm] (cond_rnd) {$t_r \sim \mathcal{U}$};

\draw[grayflow] (input.east) -| (cond_fix.south);  
\draw[grayflow] (input.east) -| (cond_rnd.north);  

\draw[grayflow] (cond_fix.east) -- (noise_fix.west);
\draw[grayflow] (cond_rnd.east) -- (noise_rnd.west);
\node[unet, right=4cm of input, xshift=18mm, yshift=0mm,
      draw=none, fill=none, drop shadow={opacity=0}] (backbone) {};
\coordinate (bb_in_fix) at (backbone.west |- noise_fix.east);
\coordinate (bb_in_rnd) at (backbone.west |- noise_rnd.east);
\coordinate (bb_out_fix) at (backbone.east |- bb_in_fix);
\coordinate (bb_out_rnd) at (backbone.east |- bb_in_rnd);
\draw[blueflow] (noise_fix.east) -- node[above, pos=0.6] {$x_{t^*}$} (bb_in_fix);
\draw[redflow]  (noise_rnd.east) -- node[below, pos=0.6] {$x_{t_{r}}$} (bb_in_rnd);
\node[tensor, anchor=west, fill=cvprblue] (COL)
  at ($(bb_out_fix)+(1.2cm,0)$) {$l^{*}$ Readout Embedding\\$e$};
\draw[blueflow] (bb_out_fix) -- (COL.west);
\node[block, right=0.8cm of COL, minimum width=1.5cm, minimum height=0.8cm] (mlp) {MLP\\$g_\phi(\cdot)$};
\node[sum, right=0.6cm of mlp] (plus) {\Large$+$};
\node[tensor, right=0.6cm of plus, fill=cvprblue!50] (z) {Clustering Embedding\\$z$};
\draw[blueflow] (COL) -- (mlp);
\draw[blueflow] (mlp) -- (plus);
\draw[blueflow] (COL.north) -- ++(0,0.35) -| node[midway, above, font=\scriptsize, color=deepblue, inner sep=1pt] {Identity} (plus.north);
\draw[blueflow] (plus) -- (z);
\node[loss, right=0.8cm of z, draw=deepblue, fill=deepblue!5] (loss_kl) {$\mathcal{L}_{\mathrm{KL}} + \mathcal{L}_{\mathrm{Gr}} + \mathcal{L}_{\mathrm{En}}$};
\draw[blueflow] (z) -- (loss_kl);
\node[tensor, anchor=west, align=center] (pred)
  at ($(bb_out_rnd)+(4.5cm,0)$) {Predicted~$\hat{\epsilon}$};
\draw[redflow] (bb_out_rnd) -- (pred.west);
\node[loss] (loss_mse)
  at (loss_kl.center |- pred.center) {$\mathcal{L}_{\mathrm{Re}}$};

\draw[redflow] (pred.east) -- coordinate(mid_loss) (loss_mse.west);

\node[above=0.45cm of mid_loss] (real_eps) {Real $\epsilon$};

\draw[redflow] (real_eps.south) -- (mid_loss);

\coordinate (total_center) at ($(loss_kl.south)!0.5!(loss_mse.north)$);
\node[tensor, font=\bfseries, anchor=center] (total)
  at (total_center) {Total Loss $\mathcal{L}$};
\draw[blueflow] (loss_kl.south) -- (total.north);
\draw[redflow]  (loss_mse.north) -- (total.south);
\begin{scope}[on background layer]
    \def\bgPadY{0.55cm}

    \coordinate (blueTL) at ($(noise_fix.north west)+(-0.2,0.3+\bgPadY)$);
    \coordinate (blueBR) at ($(loss_kl.south east)+(0.2,-0.5-\bgPadY)$);

    \fill[deepblue!3, rounded corners=8pt] (blueTL) rectangle (blueBR);
    \node[deepblue!80, font=\bfseries\scriptsize, anchor=north east]
      at ($(loss_kl.north east)+(0,0.4)$) {Branch A: Clustering Optimization};

    \coordinate (redTL) at ($(noise_rnd.north west)+(-0.2,0.3+\bgPadY)$);

    \path let \p1=(blueTL), \p2=(blueBR) in
      coordinate (redBR) at ($(redTL) + (0,{\y2-\y1})$);

    \coordinate (redBRx) at (redBR -| blueBR);

    \fill[brickred!3, rounded corners=8pt] (redTL) rectangle (redBRx);

    \node[brickred!80, font=\bfseries\scriptsize, anchor=south east]
      at ($(redBRx)+(0,-0.1)$) {Branch B: Generative Consistency};
\end{scope}

\begin{scope}[on background layer]
  \path let \p1=(blueTL), \p2=(redBRx), \p3=(backbone.center) in
    node[unet, drop shadow={opacity=3},fill=bbgrayblue,
         minimum width=3.2cm, minimum height={\y1-\y2-10}]
      (backbonevis) at (\x3,{(\y1+\y2)/2})
      {Shared Pretrained\\Diffusion U-Net\\($\epsilon_\theta$)};
\end{scope}

\begin{scope}[on background layer]
  \draw[->, gray!50, dashed, line width=1.5pt]
    (total.west) -- ++(-1.2,0) |- (backbone.east);

  \draw[->, gray!50, dashed, line width=1.5pt]
    (total.west) -- ++(-0.4,0) -| (mlp.south);
\end{scope}

\end{tikzpicture}
}
\caption{\textbf{Training framework of DiEC.}
With COL$(l^{*})$ and COT$(t^{*})$ selected, Branch A optimizes clustering using $z=e+g_{\phi}(e)$ and $\mathcal{L}_{\mathrm{KL}}+\mathcal{L}_{\mathrm{Gr}}+\mathcal{L}_{\mathrm{En}}$, while Branch B performs denoising at random $t_{\mathrm{r}}$ with $\mathcal{L}_{\mathrm{Re}}$ to preserve generative stability. Both form the total loss $\mathcal{L}$.}

\label{fig:diec_framework}
\end{figure*}

\textbf{Diffusion models.} The core idea of diffusion models is to formulate generation as iterative denoising: a forward process gradually corrupts data with Gaussian noise, and a learned reverse process progressively removes noise to generate samples from the data distribution.

In DDPM~\cite{ho2020ddpm}, the forward noising process is defined as a Markov chain that gradually corrupts a data sample $x_0 \sim q_{\mathrm{data}}(x)$:
\begin{equation}
q(x_t \mid x_{t-1}) = \mathcal{N}\!\left(x_t\middle|\, \sqrt{1-\beta_t}\,x_{t-1},\, \beta_t \mathbf{I}\right),
\end{equation}
where $\{\beta_t\}_{t=1}^T$ is a predefined noise schedule and $t\in\{1,\ldots,T\}$ denotes the diffusion timestep. 

Let $\alpha_t = 1-\beta_t$ and $\bar{\alpha}_t = \prod_{s=1}^t \alpha_s$. Then, the marginal distribution of $x_t$ admits a closed-form expression
\begin{equation}
x_t = \sqrt{\bar{\alpha}_t}\,x_0 + \sqrt{1-\bar{\alpha}_t}\,\epsilon,\qquad \epsilon \sim \mathcal{N}(0,\mathbf{I}),
\end{equation}
which shows that the noise level of $x_t$ is fully governed by the timestep $t$.

The reverse denoising process is parameterized as a Gaussian transition
\begin{equation}
p_{\theta}(x_{t-1}\mid x_t)=\mathcal{N}\!\left(x_{t-1}\middle|\,\mu_{\theta}(x_t,t),\,\sigma_t^2\mathbf{I}\right),
\end{equation}
where the mean is expressed via a noise-prediction network $\epsilon_{\theta}$ as
\begin{equation}
\mu_{\theta}(x_t,t)
=\frac{1}{\sqrt{\alpha_t}}\!\left(x_t-\frac{\beta_t}{\sqrt{1-\bar{\alpha}_t}}\,\epsilon_{\theta}(x_t,t)\right),
\end{equation}
and the variance is fixed to the posterior value as
\begin{equation}
\sigma_t^2
=\tilde{\beta}_t=\frac{1-\bar{\alpha}_{t-1}}{1-\bar{\alpha}_t}\,\beta_t .
\end{equation}
In practice, the noise prediction network $\epsilon_{\theta}(x_t,t)$ predicts the forward-process noise $\epsilon$ and is trained with a simple mean-squared error objective:
\begin{equation}
\mathcal{L}(\theta)=\mathbb{E}_{t,x_0,\epsilon}\!\left[\left\lVert \epsilon-\epsilon_{\theta}(x_t,t)\right\rVert_2^2\right].
\end{equation}

The above MSE objective can be interpreted as a simplified variational bound on the data likelihood. At the sampling stage, starting from $x_T\sim\mathcal{N}(0,\mathbf{I})$, the reverse transition $p_{\theta}(x_{t-1}\mid x_t)$ is applied to progressively denoise samples and generate data.

DDPMs use discrete timesteps, while later work introduced continuous-time variants and an SDE-based view \cite{song2021scorebased}. Recent studies have also explored Vision Transformers (ViTs) as diffusion backbones \cite{dosovitskiy2021vit,peebles2023dit}. In this paper, we follow the discrete-time DDPM formulation with a U-Net denoiser to match our timestep-selection analysis.

\paragraph{Deep Embedded Clustering.}\label{sec:background_dec}
Deep Embedded Clustering (DEC) is a method that simultaneously learns feature representations and cluster assignments~\cite{xie2016unsupervised}. Given a set of embeddings $\{z_i\}_{i=1}^{N}$ and cluster centroids $\{\mu_k\}_{k=1}^{K}$, DEC defines the soft assignment via a Student's $t$-distribution kernel as
\begin{equation}
q_{ik}=\frac{\left(1+\lVert z_i-\mu_k\rVert^2/\alpha\right)^{-\frac{\alpha+1}{2}}}{\sum_{k'=1}^{K}\left(1+\lVert z_i-\mu_{k'}\rVert^2/\alpha\right)^{-\frac{\alpha+1}{2}}}.
\label{eq:dec_q}
\end{equation}

To emphasize high-confidence assignments and mitigate cluster-size imbalance, DEC constructs a target distribution $P=[p_{ik}]$ from $Q=[q_{ik}]$ as
\begin{equation}
p_{ik}=\frac{q_{ik}^2/\sum_i q_{ik}}{\sum_{k'=1}^{K}\left(q_{ik'}^2/\sum_i q_{ik'}\right)}.
\label{eq:dec_p}
\end{equation}

DEC then minimizes the KL divergence 
\begin{equation}
\mathrm{KL}(P\|Q)=\sum_i\sum_k p_{ik}\log\frac{p_{ik}}{q_{ik}}
\label{eq:kl}
\end{equation}
for self-training, progressively sharpening the soft assignments and improving clustering consistency.

\section{Diffusion Embedded Clustering}
\begin{figure*}[t]
\centering
\resizebox{\linewidth}{!}{%
\begin{tikzpicture}

\begin{axis}[
name=hmA,
scale only axis, clip=false,
width=0.46\linewidth, height=0.26\linewidth,
axis lines=box, axis line style={black!65},
xlabel={Diffusion timestep $t$}, 
ylabel={U-Net depth}, 
xmin=-0.5,xmax=19.5, ymin=-0.5,ymax=8.5, 
xtick={0,1,2,3,4,5,6,7,8,9,10,11,12,13,14,15,16,17,18,19},
xticklabels={5,10,15,20,25,30,35,40,45,50,55,60,65,70,75,80,85,90,95,100},
xticklabel style={font=\scriptsize,rotate=45,anchor=east,inner sep=1pt},
ytick={0,1,2,3,4,5,6,7,8},
yticklabels={$D_1$,$D_2$,$D_3$,$D_4$,\textbf{$\mathrm{COL}$},$U_4$,$U_3$,$U_2$,$U_1$},
ytick pos=left,
tick align=outside, xtick pos=bottom, ytick pos=left,
tick style={black!55},
label style={font=\small}, ticklabel style={font=\scriptsize},
grid=major, grid style={black!10},
colormap={scottmap}{
  color(0cm)=(rgb,255:red,255;green,239;blue,230)
  color(1cm)=(rgb,255:red,176;green, 76;blue, 33)
},
point meta min=120000, point meta max=180000,
colorbar,
colorbar style={
  title={$\widetilde{\mathrm{SS}}_{Sm}({l,t})$},
  title style={font=\scriptsize,yshift=1pt},
  yticklabel style={
    font=\scriptsize,
    /pgf/number format/fixed,
    /pgf/number format/precision=1
  },
  y tick scale label style={font=\tiny, yshift=-2.7pt},
  width=0.18cm,
  at={(1.02,0.5)}, anchor=west,
  tick align=outside
},
]

\pgfplotstableread[col sep=space]{heatmapScott.dat}\heattableB
\addplot[matrix plot*, mesh/cols=20, point meta=explicit,
         draw=white, line width=0.18pt]
table[x=x,y=y,meta=scott]{\heattableB};

\draw[draw=deeporange!90!yellow, line width=1.6pt,
      rounded corners=1.1pt, line join=round,
      preaction={draw=white, line width=3.0pt}]
(axis cs:-0.5,3.5) rectangle (axis cs:19.5,4.5);

\end{axis}
\begin{axis}[
name=hmB,
at={(hmA.outer east)}, anchor=outer west, xshift=1.0cm, 
scale only axis, clip=false,
width=0.46\linewidth, height=0.26\linewidth,
axis lines=box, axis line style={black!65},
xlabel={Diffusion timestep $t$}, 
ylabel={}, 
xmin=-0.5,xmax=19.5, ymin=-0.5,ymax=8.5, y dir=reverse,
xtick={0,1,2,3,4,5,6,7,8,9,10,11,12,13,14,15,16,17,18,19},
xticklabels={5,10,15,20,25,30,35,40,45,50,55,60,65,70,75,80,85,90,95,100},
xticklabel style={font=\scriptsize,rotate=45,anchor=east,inner sep=1pt},
ytick={0,1,2,3,4,5,6,7,8},
yticklabels={$U_1$,$U_2$,$U_3$,$U_4$,\textbf{$\mathrm{COL}$},$D_4$,$D_3$,$D_2$,$D_1$},
tick align=outside, xtick pos=bottom, ytick pos=left,
tick style={black!55},
label style={font=\small}, ticklabel style={font=\scriptsize},
grid=major, grid style={black!10},
colormap={accmap}{
  color(0cm)=(rgb,255:red,232;green,241;blue,255)
  color(1cm)=(rgb,255:red, 23;green, 72;blue,130)
},
point meta min=0.70, point meta max=0.96,
colorbar,
colorbar style={
  title={$\mathrm{ACC}_{sm}(l,t)$},
  title style={font=\scriptsize,yshift=1pt},
  yticklabel style={font=\scriptsize},
  width=0.18cm,
  at={(1.02,0.5)}, anchor=west,
  tick align=outside
},
]

\pgfplotstableread[col sep=space]{heatmap.dat}\heattableA
\addplot[matrix plot*, mesh/cols=20, point meta=explicit,
         draw=white, line width=0.18pt]
table[x=x,y=y,meta=acc]{\heattableA};

\draw[draw=deeporange!90!yellow, line width=1.6pt,
      rounded corners=1.1pt, line join=round,
      preaction={draw=white, line width=3.0pt}]
(axis cs:-0.5,3.5) rectangle (axis cs:19.5,4.5);

\end{axis}

\end{tikzpicture}%
}
\caption{\textbf{Selection of COL on USPS.} The heatmaps display metrics across layers and timesteps. Left: Smoothed Scott Score ($\widetilde{\mathrm{SS}}_{Sm}$). Right: ground-truth Smoothed Accuracy Score($\mathrm{ACC}_{Sm}$). The consistent alignment between the two metrics validates the COL selected by $\widetilde{\mathrm{SS}}_{Sm}$. On this dataset, it is located at the bottleneck layer.}
\label{fig:heatmap_layer_timestep_usps_acc_scott}
\end{figure*}
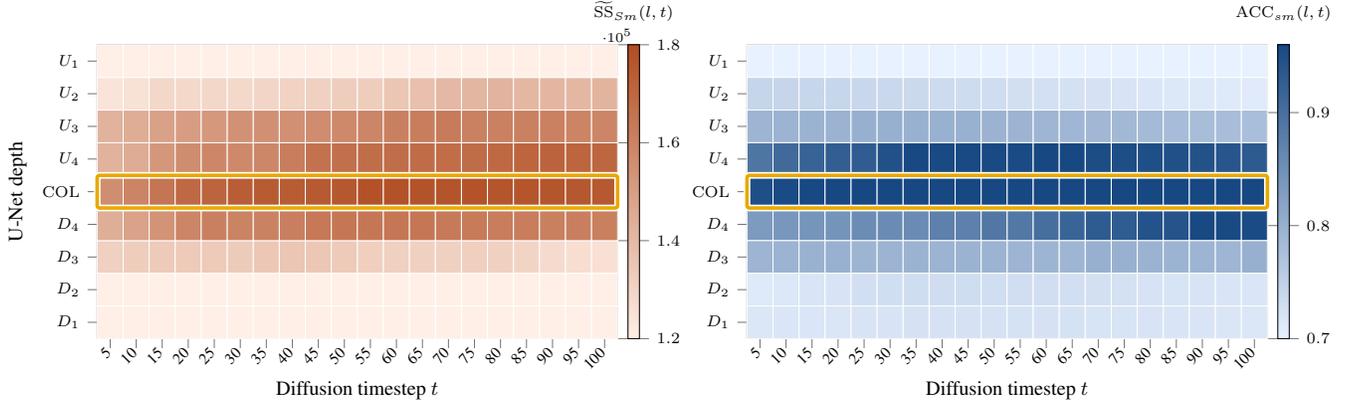
To leverage the multi-scale representations of diffusion models for clustering, DiEC first identifies the COL + COT by searching the layer$\times$timestep space of pretrained diffusion models, which guide the extraction, as illustrated in Fig.~\ref{fig:teaser}. Then, following Fig.~\ref{fig:diec_framework}, DiEC learns the embedding representations at the selected COL + COT through a shared pretrained U-Net backbone, and decouples the clustering representations via a lightweight residual network. Finally, DiEC performs clustering based on a DEC-style KL-divergence objective, supplemented by a random-timestep diffusion denoising objective to maintain the generative capability of pretrained diffusion models.

\subsection{Locating Clustering-Friendly Diffusion Representations}
\begin{algorithm}[tb]
\caption{DiEC Joint Training Procedure}
\label{alg:diec_training}
\textbf{Input}: Unlabeled dataset $\mathcal{D}$, Pretrained U-Net $\epsilon_{\theta}$ with COL $l^*$, COT $t^*$, Residual MLP $g_{\phi}$, Cluster number $K$, MaxEpochs $E_{\max}$, Update interval $I_{\mathrm{target}}$. \\
\begin{algorithmic}[1]
\STATE \textbf{// Phase 1: Robust Initialization}
\STATE Extract averaged embeddings $\{\bar{z}_i\}$ for all $x_i \in \mathcal{D}$ at selected ($l^*$, $t^*$) using $M$ noise trials.
\STATE Initialize centroids $\{\mu_k\}_{k=1}^K$ via $k$-means on $\{\bar{z}_i\}$.
\STATE Construct graph affinity matrix $\mathbf{S}$ (Eq.~\ref{eq:sij})
\vspace{0.1cm}
\STATE \textbf{// Phase 2: Joint Optimization}
\FOR{epoch $= 1$ to $E_{\max}$}
    \IF{epoch $\%$ $I_{\mathrm{target}} = 0$}
        \STATE Update target distribution $P$ (Eq.~\ref{eq:dec_p}). 
    \ENDIF
    \FOR{each sampled from $\mathcal{D}$}
        \STATE \textbf{// Branch A: Clustering-Aware Optimization} 
        \STATE Sample noise $\epsilon \sim \mathcal{N}(0, \mathbf{I})$ and generate $x_{t^*, i}$.
        \STATE Extract features $e_i = h_{\theta}^{l^*}(x_{t^*, i}, t^*)$.
        \STATE Compute residual embedding (Eq.~\ref{eq:z}).
        \STATE Compute soft assignments $Q$ (Eq.~\ref{eq:dec_q}) and clustering loss $\mathcal{L}_{\mathrm{KL}}$ (Eq.~\ref{eq:kl}).
        \STATE Compute graph constraint loss $\mathcal{L}_{\mathrm{Gr}}$, $\mathcal{L}_{\mathrm{En}}$  (Eq~\ref{eq:lgr}, ~\ref{eq:len}).

        \vspace{0.1cm}
        \STATE \textbf{// Branch B: Diffusion Consistency}
        \STATE Sample random timesteps $t_{\mathrm{r}} \sim \mathcal{U}(\{1,\dots,T\})$.
        \STATE Generate noisy inputs $x_{t_{\mathrm{r}}, i}$.
        \STATE Compute denoising loss $\mathcal{L}_{\mathrm{Re}}$ (Eq.~\ref{eq:lre}).
        \vspace{0.1cm}
        \STATE \textbf{// Joint Update}
        \STATE Update $\mathbf{S}$ via Lagrange multipliers.
        \STATE $\mathcal{L}_{\mathrm{total}} = \mathcal{L}_{\mathrm{Re}} + \alpha\mathcal{L}_{\mathrm{KL}} + \beta\mathcal{L}_{\mathrm{Gr}} + \gamma\mathcal{L}_{\mathrm{En}}$.
        \STATE Update $\theta, \phi, \{\mu_k\}$ via gradient descent: $\nabla \mathcal{L}_{\mathrm{total}}$.
    \ENDFOR
\ENDFOR 
\STATE \textbf{Output:} Cluster centroids $\{\mu_k\}$, assignments $Q$.
\end{algorithmic}
\end{algorithm}
\paragraph{Unsupervised Clusterability Metric.}
In the U-Net-based diffusion models, multi-scale semantic representations are indexed by both the network layer $l$ and the noise timestep $t$. Given an unlabeled dataset $\mathcal{D} = \{x_{0,i}\}_{i=1}^{N}$ and the number of clusters $K$, $x_{t,i}$ denotes the noisy observation of the clean sample $x_{0,i}$ at diffusion timestep $t$. Then, we can obtain the embedding representations $\mathbf{E}_{l,t} = \{{e}_i^{\, l}(t)\}_{i=1}^{N}$ on U-Net layer $l$ at timestep $t$ as
\begin{equation}
e_i^{\,l}(t) = h_{\theta}^{l}(x_{t,i}, t).
\end{equation}

The clusterability of these representations can be evaluated using the Scott Score, denoted as $\mathrm{SS}(l,t)$, an unsupervised metric applied to assess clustering performance on data without ground-truth labels.

We run $k$-means on $\mathbf{E}_{l,t} = \{e_i^{\,l}(t)\}_{i=1}^{N}$ to yield
cluster assignments $C_k^{\,l}(t)$ and centroids $\mu_k^{\,l}(t)$, and define the within-cluster scatter and the between-cluster scatter as
\begin{equation}
\mathbf{W}^{\,l}(t) = \sum_{k=1}^{K} \sum_{i \in C_k^{\,l}(t)} \bigl( e_i^{\,l}(t) - \mu_k^{\,l}(t) \bigr) \bigl( e_i^{\,l}(t) - \mu_k^{\,l}(t) \bigr)^{\top},
\end{equation}
\begin{equation}
\mathbf{B}^{\,l}(t) = \sum_{k=1}^{K} \left| C_k^{\,l}(t) \right| \bigl( \mu_k^{\,l}(t) - \bar{e}^{\,l}(t) \bigr) \bigl( \mu_k^{\,l}(t) - \bar{e}^{\,l}(t) \bigr)^{\top},
\end{equation}
where $\bar{e}^{\,l}(t) = \frac{1}{N} \sum_{i=1}^{N} e_i^{\,l}(t)$ denotes the global mean embedding across all samples.

Then, the Scott Score is defined as
\begin{equation}
\mathrm{SS}(l,t) = N \Bigl( \log \det \mathbf{T}^{\,l}(t) - \log \det \mathbf{W}^{\,l}(t) \Bigr),
\end{equation}
where $\mathbf{T}^{\,l}(t) = \mathbf{W}^{\,l}(t) + \mathbf{B}^{\,l}(t)$.

To reduce noise, we smooth the Scott Score along timesteps by a centered moving average as 
\begin{equation}
\mathrm{SS}_{Sm}(l,t) = \frac{1}{|\mathcal{N}(t)|} \sum_{\tau \in \mathcal{N}(t)} \mathrm{SS}(l,\tau),
\end{equation}
where $\mathcal{N}(t)=\left\{\,\tau\in\mathcal{T}\;\Big|\;|\tau-t|\le\lfloor w/2\rfloor\,\right\}$, $\mathcal{T}$ is the set of evaluated timesteps and $w$ is the window size. 

In DiEC, this smoothed Scott Score $\mathrm{SS}_{Sm}(l,t)$ serves as the unsupervised internal criterion for searching clustering-friendly representations.

\paragraph{Optimal Search for COL and COT.}
Based on the smoothed Scott Score, we design an Optimal Search strategy for identifying the COL and COT, denoted as $l^*$ and $t^*$ respectively. This strategy consists of two sequential stages: layer selection, followed by timestep selection. 

\textbf{In layer selection}, considering the dimensional differences between representations across layers in the U-Net, we apply a PCA-based layer-wise alignment map $\mathcal{A}_l(\cdot)$ to obtain aligned embeddings $\widetilde{\mathbf{E}}_{l,t}=\{\tilde{e}_i^{\,l}(t)\}_{i=1}^{N}$, where $\tilde{e}_i^{\,l}(t)=\mathcal{A}_l({e}_i^{\,l}(t))$, and evaluate its smoothed Scott Score as $\widetilde{\mathrm{SS}}_{Sm}(l,t)$.

To obtain the robust evaluation for each layer, the timesteps ranked in the top-$\rho$ percent (e.g. $\rho$=20) are utilized to identify the COL as
\begin{equation}
l^*=\arg\max_{l} \frac{1}{|\mathcal{T}_l^{\rho}|}\sum_{t\in \mathcal{T}_l^{\rho}} \widetilde{\mathrm{SS}}_{Sm}(l,t),
\end{equation}
where $\mathcal{T}_l^{\rho}
=\Bigl\{\, t\in\mathcal{T}\ \Big|\ 
\mathrm{rank}_t\!\big(\widetilde{\mathrm{SS}}_{Sm}(l,t)\big)\le \lceil \rho|\mathcal{T}|\rceil \Bigr\}$.

From Fig.~\ref{fig:heatmap_layer_timestep_usps_acc_scott}, we can see that the COL is located at the bottleneck layer on the USPS dataset (left panel). At the COL, the unsupervised Scott Score (SS) computed from the PCA-aligned inputs and the ground-truth clustering accuracy (ACC) exhibit similar trends. This observation confirms the effectiveness of the smoothed Scott Score as a reliable metric for identifying the COL.
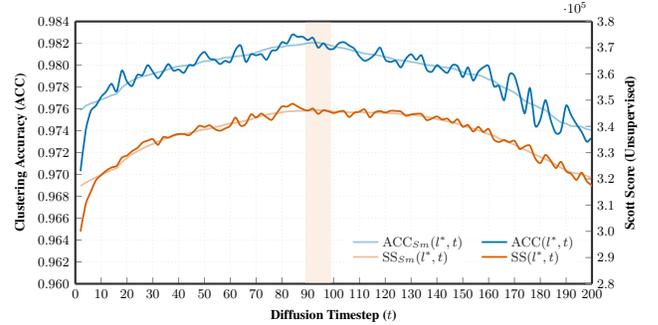
\begin{figure}[t]
  \centering
  \resizebox{0.98\columnwidth}{!}{%
    \begin{tikzpicture}
      \begin{axis}[
        width=12.2cm, height=6.2cm,
        scale only axis,
        xmin=0, xmax=200,
        ymin=0.960, ymax=0.984,
        axis y line*=left,
        xtick pos=bottom,
        xlabel={Diffusion Timestep ($t$)},
        ylabel={Clustering Accuracy (ACC)},
        label style={font=\small\bfseries},
        xticklabel style={black, font=\footnotesize},
        xlabel style={black, font=\small\bfseries},
        ylabel style={black},
        yticklabel style={black, font=\footnotesize,/pgf/number format/precision=3,/pgf/number format/fixed zerofill},
        grid=major,
        grid style={dotted, softGrid, line width=0.8pt},
        axis line style={black!80},
        line width=1.45pt,
        xtick={0,10,...,200},
        ytick={0.960,0.962,...,0.984}
      ]

        \addplot[
          draw=none,
          fill=sciRed!10,
          forget plot
        ] coordinates {(89,0.960099) (99,0.960099) (99,0.983901) (89,0.983901)} \closedcycle;
        \addplot[sciBlue, very thick, smooth] table[x=t, y=acc] {plotdata.dat};
        \label{plot:acc}
        \addplot[sciBlue, very thick,opacity=0.4] table[x=t, y=acc_smoothed] {plotdata.dat};
        \label{plot:sacc}
      \end{axis}

      \begin{axis}[
        width=12.2cm, height=6.2cm,
        scale only axis,
        xmin=0, xmax=200,
        ymin=280000, ymax=380000,
        axis y line*=right,
        axis x line=none,
        ylabel={Scott Score (Unsupervised)},
        label style={font=\small\bfseries},
        ylabel style={black},
        axis line style={black!80, line width=1.45pt},
        scaled y ticks=true,
        ytick={280000,290000,...,380000},
        yticklabel style={
          black, font=\footnotesize,
          /pgf/number format/fixed,
          /pgf/number format/fixed zerofill,
          /pgf/number format/precision=1
        }
      ]
        \addplot[sciRed, very thick, smooth] table[x=t, y=Scott] {plotdata.dat};
        \label{plot:scott}
        \addplot[sciRed, very thick, opacity=0.4] table[x=t, y=Scott_Smoothed] {plotdata.dat};
        \label{plot:sscott}
        \node[draw=none, fill=white, fill opacity=0.9, rounded corners=2pt, anchor=south east, font=\footnotesize]
        at (axis cs:195, 285000) {
        \begin{tabular}{@{}l@{ }l@{}}
          \ref{plot:acc}   & \textcolor{black}{\textbf{${\mathrm{ACC}}(l^*,t)$}} \\
          \ref{plot:scott} & \textcolor{black}{\textbf{${\mathrm{SS}}(l^*,t)$}}
        \end{tabular}
        };
        \node[draw=none, fill=white, fill opacity=0.9, rounded corners=2pt, anchor=south east, font=\footnotesize]
        at (axis cs:152, 285000) {
        \begin{tabular}{@{}l@{ }l@{}}
          \ref{plot:sacc}   & \textcolor{black}{\textbf{${{\mathrm{ACC}_{Sm}}(l^*,t)}$}} \\
          \ref{plot:sscott} & \textcolor{black}{\textbf{${{\mathrm{SS}_{Sm}}(l^*,t)}$}}
        \end{tabular}
        };
      \end{axis}
    \end{tikzpicture}%
  }
  \vspace{-2mm}
    \caption{\textbf{Selection of COT on MNIST.} The evolution of the unsupervised Scott Score ($\mathrm{SS}$ and $\mathrm{SS}_{sm}$, orange) is closely aligned with the ground-truth clustering accuracy ($\mathrm{ACC}$ and $\mathrm{ACC}_{sm}$, blue) across diffusion timesteps on a fixed COL. Both metrics exhibit a distinct unimodal trend, with their smoothed peaks aligning in the same interval ($t \in [89, 99]$).}
  \label{fig:ablation_acc_scott_advanced}
\end{figure}

\textbf{In timestep selection}, with $l^{*}$ fixed, we evaluate clusterability directly on the native representations to remain consistent with training. The COT is then identified by
\begin{equation}
t^{*}=\arg\max_{t} \mathrm{SS}_{Sm}(l^*,t).
\end{equation}

Fig.~\ref{fig:ablation_acc_scott_advanced} shows the trajectory alignment between the unsupervised Scott Score (SS) and ground-truth clustering accuracy (ACC) across diffusion timesteps on the MNIST dataset. We also plot their smoothed values using the centered moving average mentioned above. It can be clearly seen that SS exhibits the same trend as ACC. In particular, by removing the effect of noise, their smoothed values are perfectly matched, all reaching the maximum at the same timestep. This demonstrates that the smoothed Scott Score is a suitable evaluation metric for locating COT.

The complete procedure is detailed in Section~\ref{app:A} of the Appendix. Specifically, to mitigate diffusion stochasticity and reduce computational overhead, we estimate scores on a sample subset and average them over several independent noise realizations. This strategy substantially accelerates inference while maintaining high clustering accuracy.

\subsection{Clustering-Aware Optimization}
Given the features $e_i^{\,l^*}(t^{*})$ extracted at $(l^{*},t^{*})$, we obtain clustering representations $z_i$ through a lightweight residual mapping, and optimize a DEC-style KL self-training objective with graph regularization to strengthen cluster structure. Meanwhile, we add a standard denoising loss at a random timestep $t_r$, which helps maintain diffusion consistency and stabilize the representations.

\paragraph{Residual-decoupled embedding.}
To simplify the notation, we refer to $e_i^{\,l^*}(t^{*})$ as $e_i$, where
\begin{equation}
e_i = h_{\theta}^{l^{*}}(x_{t^{*}, i}, t^{*}).
\end{equation}

To reduce potential interference between the denoising objective and the clustering objective, we adopt a lightweight residual mapping $g_{\phi}(\cdot)$ (a two-layer ReLU MLP with matched input--output dimensions) to decouple clustering adaptation from the pretrained representation. The final clustering embedding is defined as
\begin{equation}
\label{eq:z}
z_i = e_i + g_{\phi}(e_i).
\end{equation}

This residual branch partially decouples clustering from denoising and enhances the expressiveness of the clustering loss.
\paragraph{Clustering objective.}
Given the clustering embeddings $\{z_i\}_{i=1}^{N}$, we adopt a DEC-style KL-divergence self-training objective. Since diffusion noising introduces stochasticity that can affect centroid initialization, we compute the mean embedding by averaging over $M$ independent noise realizations at the selected layer $l^{*}$ and timestep $t^{*}$ as
\begin{equation}
\label{eq:z_mean}
\bar{z}_i=\frac{1}{M}\sum_{m=1}^{M} z_i^{(m)}.
\end{equation}

Then, $k$-means is applied to $\{\bar{z}_i\}_{i=1}^{N}$ to initialize the centroids $\{\mu_k\}_{k=1}^{K}$. Based on $\{z_i\}_{i=1}^{N}$ and $\{\mu_k\}_{k=1}^{K}$, we compute the soft assignments $Q$ via Eq.~\eqref{eq:dec_q} and the target distribution $P$ via Eq.~\eqref{eq:dec_p}. The clustering objective is then defined as their KL divergence in Eq.~\eqref{eq:kl}.

\setcounter{table}{1}
\begin{table*}[t]
    \caption{\textbf{Quantitative comparison with state-of-the-art methods.} 
    We primarily benchmark against \textbf{embedding-based} approaches for comparison, while also including \textbf{augmentation-consistency} methods to demonstrate the excellent clustering performance of our model. We report \textbf{ACC}, \textbf{NMI}, and \textbf{ARI} (\%).}
    \label{tab:main_acc_nmi_ari}
    \centering
    \small
    \setlength{\tabcolsep}{3.2pt}
    \renewcommand{\arraystretch}{1.08}
    \rowcolors{5}{rowgray}{white}

    \begin{tabular}{l c c c c c c c c c c c c}
        \toprule
        \rowcolor{white}
        \multirow{2}{*}{\textbf{Method}} &
        \multicolumn{3}{c}{\textbf{MNIST}} &
        \multicolumn{3}{c}{\textbf{USPS}} &
        \multicolumn{3}{c}{\textbf{F-MNIST}} &
        \multicolumn{3}{c}{\textbf{CIFAR-10}} \\
        \cmidrule(lr){2-4}\cmidrule(lr){5-7}\cmidrule(lr){8-10}\cmidrule(lr){11-13}
        \rowcolor{white}
        & \textbf{ACC} & \textbf{NMI} & \textbf{ARI}
        & \textbf{ACC} & \textbf{NMI} & \textbf{ARI}
        & \textbf{ACC} & \textbf{NMI} & \textbf{ARI}
        & \textbf{ACC} & \textbf{NMI} & \textbf{ARI} \\
        \midrule

        \rowcolor{white}
        \multicolumn{13}{l}{\textit{\textbf{Embedding-based Deep Clustering}}} \\
        DEC        & 84.7 & 79.1  & 74.9  & 73.3 & 70.6  & 63.7  & 53.6 & 59.1 & 42.0  & 25.1 & 13.6  & 7.8 \\
        IDEC       & 88.1 & 86.7  & 85.0  & 76.2 & 75.6  & 67.9  & 55.2   & 60.4  & 44.2  & 25.3   & 13.8  & 8.1 \\
        DCEC       & 88.5 & 87.6  & 84.3  & 78.0 & 81.8  & 73.6  & 62.7 & 65.9 & 54.0 & 28.7  & 26.5  & 16.9 \\
        VaDE       & 94.3 & 86.6  & 85.1  & 83.3 & 83.3  & 77.7  & 62.9 & 61.1  & 53.3  & 22.8 & 10.9  & 6.5 \\
        ClusterGAN & 95.0 & 89.0  & 89.0  & 80.2   & 75.6  & 72.0  & 63.0 & 64.0  & 50.0  & 26.2 & 23.3  & 16.8 \\
        ClusterDDPM& 97.7 & 94.0 & 93.2  & 81.3  & 83.8   & 76.1 & 70.5 & 66.1  & 56.9  & 30.5 & 17.3  & 15.7 \\
        \arrayrulecolor{rulecolor}\midrule\arrayrulecolor{black}

        \rowcolor{white}
        \multicolumn{13}{l}{\textit{\textbf{Reference: Augmentation Consistency Methods}}} \\
        IIC     & 99.2   & 97.9  & 97.8  & 81.8   & 85.9  & 81.1  & 65.7   & 63.4  & 52.4  & 61.7   & 51.3  & 41.1\\
        DCCM       & 98.2   & 95.1  & 95.1  & 86.9   & 88.6  & 83.4  & 75.3   & 68.4  & 60.2  & 62.3  & 49.6  & 40.8 \\
        CC     & 96.6   & 93.2  & 93.1  & 91.5   & 90.0  & 83.2  & 70.8   & 67.5  & 56.5  & \textbf{79.0}  & \textbf{70.5}  & \textbf{63.7} \\
        \midrule
        \hiderowcolors
        \textbf{DiEC (Ours)} &
        \textbf{99.5}  & \textbf{98.5} & \textbf{98.0} &
        \textbf{98.5} & \textbf{97.0} & \textbf{95.7} &
        \textbf{79.7}  & \textbf{75.5} & \textbf{66.5} &
        \textbf{71.4} & \textbf{61.5} & \textbf{56.1} \\
        \bottomrule
    \end{tabular}
\end{table*}

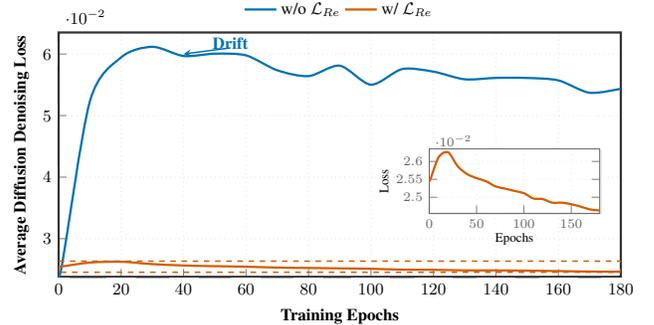
\begin{figure}[!b]
  \centering
  \resizebox{0.98\columnwidth}{!}{%
  \begin{tikzpicture}
    \pgfplotsset{compat=1.18}
    
\pgfplotstableread[row sep=\\, col sep=space]{ epoch withRe withoutRe \\ 0 0.02548702497 0.02548702497 \\ 1 0.02548702497 0.02548702497 \\ 10 0.02612113528 0.05230867833 \\ 20 0.02625464283 0.05945392572 \\ 30 0.02586529280 0.06117542619 \\ 40 0.02564684671 0.05968149665 \\ 50 0.02553267182 0.06006031044 \\ 60 0.02544660453 0.05978190895 \\ 70 0.02530243273 0.05735887295 \\ 80 0.02524170257 0.05641615802 \\ 90 0.02517566309 0.05810737240 \\ 100 0.02511409328 0.05502129709 \\ 110 0.02496518194 0.05755868659 \\ 120 0.02494854699 0.05713750655 \\ 130 0.02484785532 0.05589475840 \\ 140 0.02484911642 0.05612403125 \\ 150 0.02480497570 0.05611305306 \\ 160 0.02473801321 0.05571855114 \\ 170 0.02465863923 0.05370883705 \\ 180 0.02463173850 0.05437249026 \\ }\datatable

    \begin{axis}[
      width=12.2cm, height=6.2cm,
      xlabel={Training Epochs},
      ylabel={Average Diffusion Denoising Loss},
      xmin=0, xmax=180,
      ymin=0.0238, ymax=0.0635,
      axis line style={black!80},  
      grid=major,
      xtick pos=bottom,
      ytick pos=bottom,
      grid style={dotted, softGrid, line width=0.8pt}, 
      ticklabel style={font=\footnotesize, fill=white}, 
      label style={font=\small\bfseries}, 
      legend style={
        at={(0.5,1.03)}, anchor=south, legend columns=-1,
        draw=none, fill=none, font=\small
      },
      line width=1.2pt, 
    ]

      \addplot[sciBlue, smooth] table[x=epoch,y=withoutRe] {\datatable};
      \addlegendentry{w/o $\mathcal{L}_{Re}$}

      \addplot[sciRed, smooth] table[x=epoch,y=withRe] {\datatable};
      \addlegendentry{w/ $\mathcal{L}_{Re}$}

      \node[font=\footnotesize\bfseries, text=sciBlue] at (axis cs:55,0.0618) {Drift};
      \draw[->, >=stealth, thick, sciBlue] (axis cs:55,0.061) -- (axis cs:40,0.060);

      \draw[sciRed, thick, dashed] (axis cs:0,0.02455) rectangle (axis cs:180,0.02635);
    \end{axis}

    \begin{axis}[
      at={(7.0cm,1.2cm)}, anchor=south west,
      width=4.8cm, height=2.8cm,
      xmin=0, xmax=180,
      ymin=0.02455, ymax=0.02635,
      axis background/.style={fill=white, drop shadow={opacity=0.15,shadow xshift=1pt,shadow yshift=-1pt}},
      grid=major,
      grid style={dotted, softGrid},
      ticklabel style={font=\scriptsize, color=sciGray},
      xlabel style={font=\scriptsize, at={(0.5,-0.2)}, anchor=north},
      ylabel style={font=\scriptsize, at={(-0.2,0.5)}, anchor=south},
      xlabel={Epochs},
      ylabel={Loss},
      xtick pos=bottom,
      axis line style={black!60},
    ]
      \addplot[sciRed, very thick, smooth] table[x=epoch,y=withRe] {\datatable};
    \end{axis}
  \end{tikzpicture}%
  }
  \vspace{-2mm}
    \caption{\textbf{Denoising stability analysis on USPS.} Without $\mathcal{L}_{Re}$ (blue), the average diffusion denoising loss drifts, which impairs the generative capability of diffusion models. With $\mathcal{L}_{Re}$ (red), the model maintains a low average denoising loss.}
  \label{fig:denoise_stability_usps}
\end{figure}
\paragraph{Adaptive graph regularization.}\label{sec:graph_reg}
To preserve local consistency in the embedding space, we introduce an adaptive graph regularization term. Specifically,  we construct a row-normalized affinity matrix $\mathbf{S}=[s_{ij}]$ based on the $k$-NN neighborhood $\mathrm{NB}_i$ of each sample, with
\begin{equation}
\label{eq:sij}
s_{ij} = 
\begin{cases} 
\exp\left(-\frac{\|x_i - x_j\|_2^2}{\sigma}\right), & \text{if } x_j \in \text{NB}_i \\
0, & \text{if } x_j \notin \text{NB}_i
\end{cases}.
\end{equation}

We impose a graph constraint on the soft assignments $Q$ to strengthen the cluster structure as
\begin{equation}
\label{eq:lgr}
\begin{gathered}
\mathcal{L}_{\mathrm{Gr}}
=\sum_{i=1}^{N}\sum_{j\in \mathrm{NB}_i}\sum_{k=1}^{K}
s_{ij}\bigl(q_{ik}-q_{jk}\bigr)^2,\\
\text{s.t. }\quad
\sum_{j\in \mathrm{NB}_i} s_{ij}=1,\qquad s_{ij}\in[0,1].
\end{gathered}
\end{equation}

In addition, we introduce an entropy regularizer to prevent trivial solutions, formulated as
\begin{equation}
\label{eq:len}
\mathcal{L}_{\mathrm{En}}
=\sum_{i=1}^{N}\sum_{j\in \mathrm{NB}_i} s_{ij}\log s_{ij}.
\end{equation}

\paragraph{Joint training with diffusion denoising.}\label{sec:diff_consistency}
The update of the shared U-Net network only at timestep $t^{*}$ may weaken its denoising performance at other timesteps,
thereby impairing the overall stability of diffusion representations. Thus we introduce an additional noise-reconstruction branch at random timesteps $t_r$ to maintain generative capability. Specifically, we sample $t_r\sim \mathcal{U}(\{1,\ldots,T\})$ in each iteration and apply the standard forward noising to obtain $x_{t_r,i}$, and the noise-prediction MSE objective is then utilized to preserve the model’s generative capability, as defined by
\begin{equation}
\label{eq:lre}
\mathcal{L}_{\mathrm{Re}}
=\mathbb{E}_{t_r,\epsilon}\!\left[
\frac{1}{N}\sum_{i=1}^{N}
\left\lVert \epsilon-\epsilon_{\theta}\!\left(x_{t_r,i},t_r\right)\right\rVert_2^{2}
\right].
\end{equation}
As shown in Fig.~\ref{fig:denoise_stability_usps}, $\mathcal{L}_{\mathrm{Re}}$ prevents drift during optimization and helps preserve generative capability across timesteps. Furthermore, Appendix Section~\ref{app:B} provides qualitative results that confirm the maintained generative quality. 

\subsection{Loss Function}
\label{sec:loss}
We optimize the diffusion U-Net and the residual MLP module end-to-end with a weighted combination of objectives. The overall training objective is
\begin{equation}
\mathcal{L}
=\mathcal{L}_{\mathrm{Re}}
+\alpha\,\mathcal{L}_{\mathrm{KL}}
+\beta\,\mathcal{L}_{\mathrm{Gr}}
+\gamma\,\mathcal{L}_{\mathrm{En}}
,
\end{equation}
where $\mathcal{L}_{\mathrm{Re}}$ is the diffusion-consistency denoising reconstruction loss at random timesteps, $\mathcal{L}_{\mathrm{KL}}$ is the DEC-style clustering loss, $\mathcal{L}_{\mathrm{Gr}}$ and $\mathcal{L}_{\mathrm{En}}$ are the adaptive regularization terms. $\alpha$, $\beta$, and $\gamma$ are weighting hyperparameters that balance the relative contributions of the loss terms. The complete training procedure is provided in Algorithm~\ref{alg:diec_training}.

\begin{figure*}[!t]
  \centering
  \setlength{\tabcolsep}{2pt}

  \begin{minipage}[t]{0.245\textwidth}
    \centering
    \includegraphics[width=\linewidth]{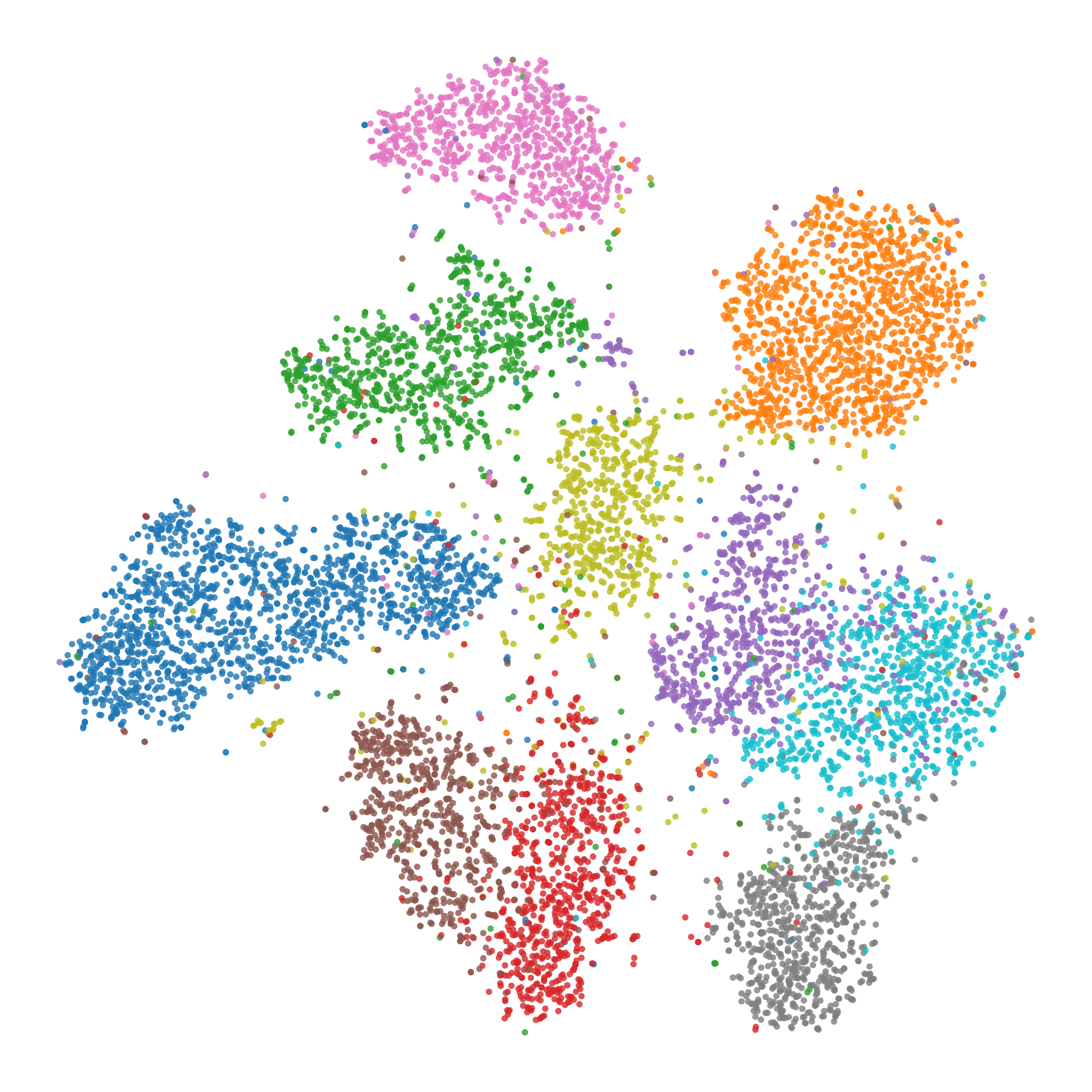}
    \vspace{-2mm}
    {\small (a) 10 epoch, t{=}1}
  \end{minipage}\hfill
  \begin{minipage}[t]{0.245\textwidth}
    \centering
    \includegraphics[width=\linewidth]{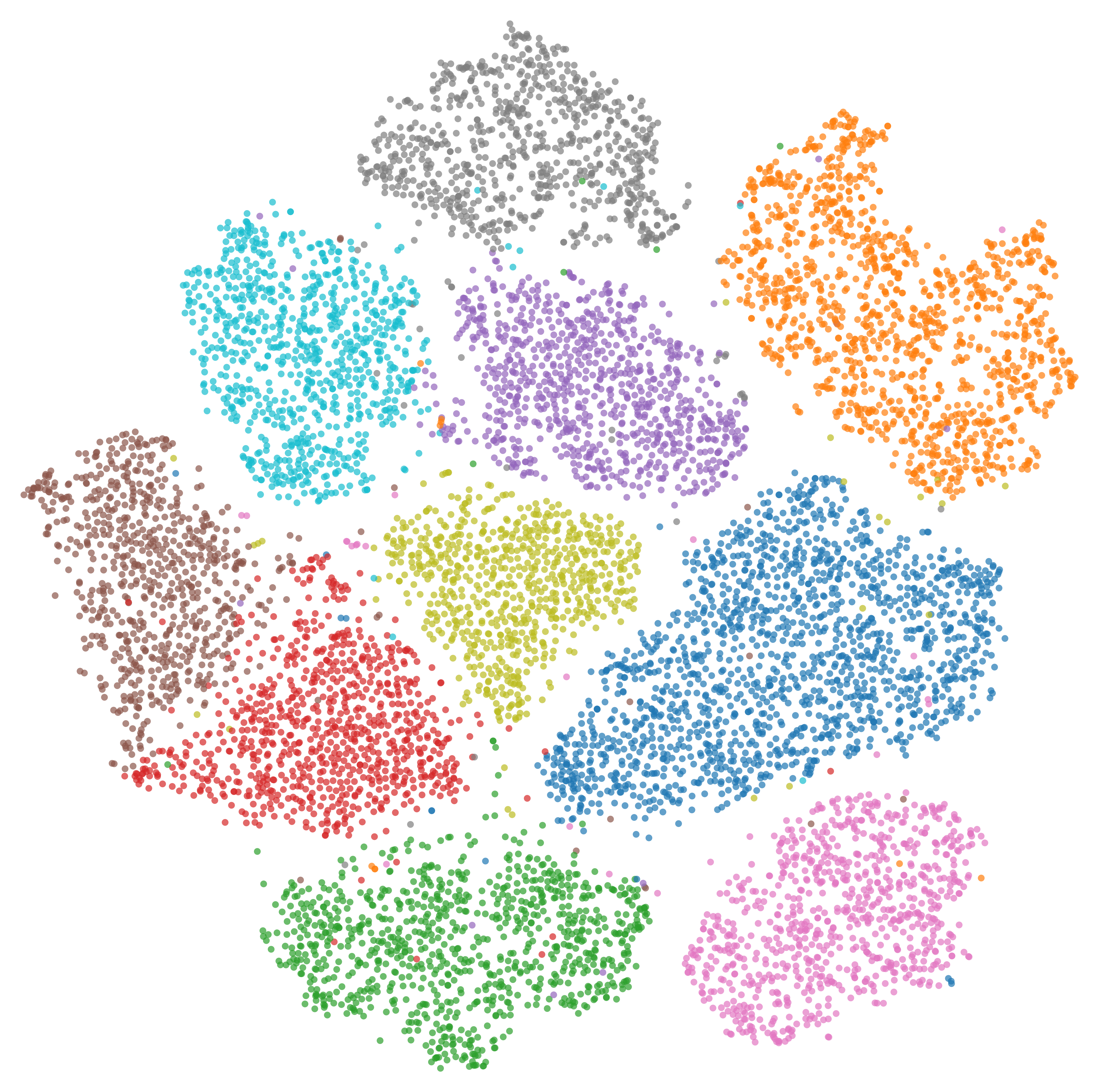}
    \vspace{-2mm}
    {\small (b) 400 epoch, t{=}1}
  \end{minipage}\hfill
  \begin{minipage}[t]{0.245\textwidth}
    \centering
    \includegraphics[width=\linewidth]{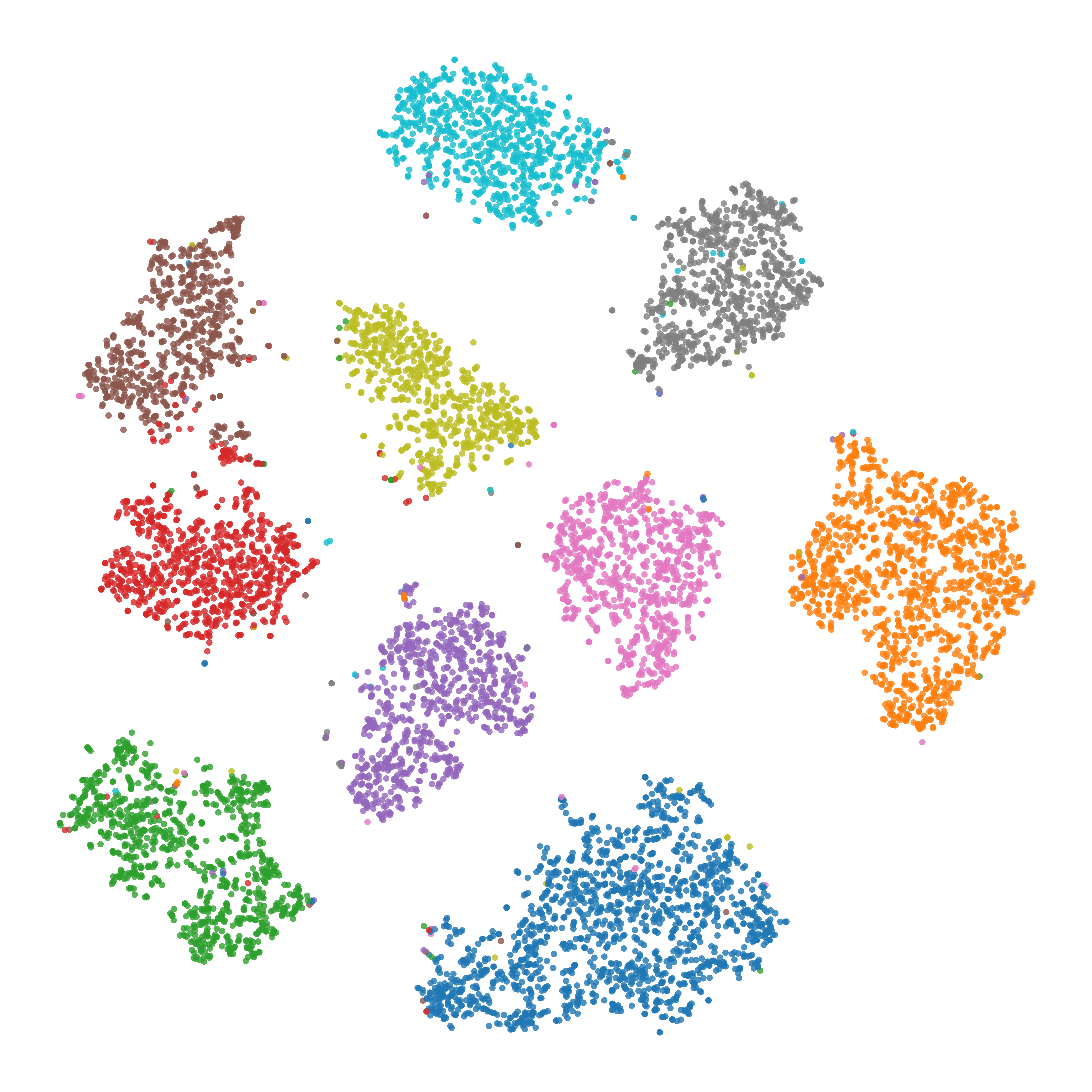}
    \vspace{-2mm}
    {\small (c) 400 epoch, t{=}$t^*$}
  \end{minipage}\hfill
  \begin{minipage}[t]{0.245\textwidth}
    \centering
    \includegraphics[width=\linewidth]{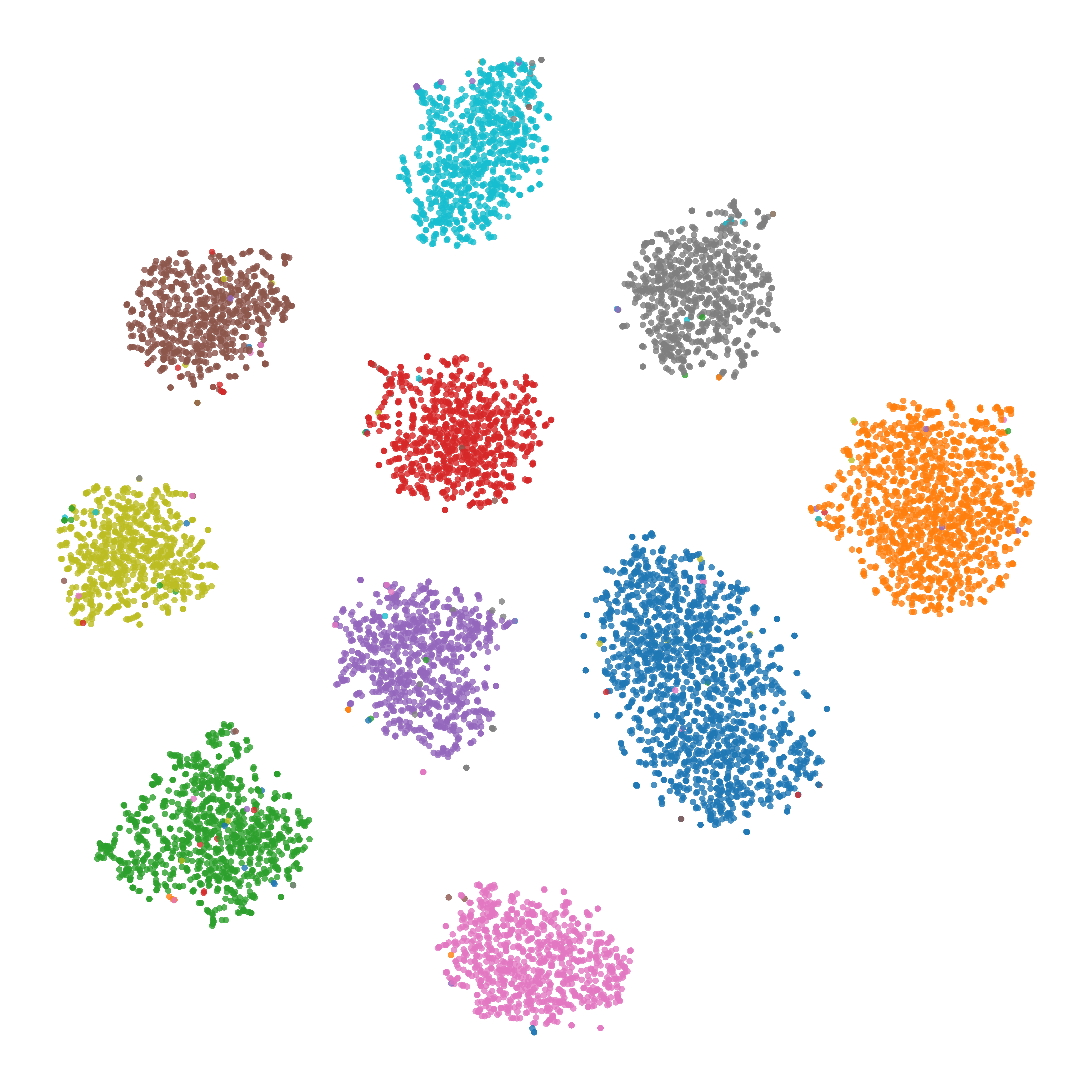}
    \vspace{-2mm}
    {\small (d) 400 epoch, t{=}$t^*$, fine-tuned}
  \end{minipage}

\caption{\textbf{Evolution of diffusion features for clustering on USPS.} 
t-SNE of COL embeddings across different pretraining stages and timesteps. We observe that clusters become more separable as diffusion pretraining progresses. Furthermore, switching from $t{=}1$ to the optimal timestep $t^*$ reveals significantly clearer structures, which are finally sharpened by DiEC optimization.}
  \label{fig:usps_tsne_1x4}
\end{figure*}

\section{Experiments}\label{sec:experiments}
\paragraph{Datasets.}
We evaluate DiEC on four widely used benchmarks: MNIST, USPS, Fashion-MNIST, and CIFAR-10. MNIST and Fashion-MNIST each comprise 70,000 grayscale images at $28\times 28$ resolution. USPS contains 9,298 grayscale images at $16\times 16$ resolution, and CIFAR-10 consists of 60,000 color images at $32\times 32$ resolution. The dataset statistics are summarized in Table~\ref{tab:datasets}.

\paragraph{Evaluation metrics.}
We evaluate clustering performance using clustering accuracy (ACC), normalized mutual information (NMI), and adjusted Rand index (ARI). Higher values indicate better clustering quality.
\setcounter{table}{0}
\begin{table}[H]
    \centering
    \small
    \rowcolors{0}{rowgray}{white}
    \caption{Summary of datasets used in our experiments.}
    \label{tab:datasets}
    \begin{tabular}{c c c c}
        \toprule
        \rowcolor{white}
        \textbf{Dataset} & \textbf{Samples} & \textbf{Resolution} & \textbf{Classes} \\
        \midrule
        MNIST         & 70{,}000 & $28\times 28$ & 10 \\
        USPS          & 9{,}298  & $16\times 16$ & 10 \\
        Fashion-MNIST & 70{,}000 & $28\times 28$ & 10 \\
        CIFAR-10      & 60{,}000 & $32\times 32$ & 10 \\

        \bottomrule
    \end{tabular}
\end{table}

\paragraph{Experimental Setup.}
We use a pretrained DDPM U-Net as the backbone. Table~\ref{tab:main_acc_nmi_ari} compares DiEC with embedding-based baselines, including DEC, IDEC, DCEC, VaDE, ClusterGAN, and ClusterDDPM. To demonstrate the competitive advantages of our approach, we also compare with augmentation-consistency methods such as IIC \cite{ji2019invariant}, DCCM \cite{wu2019deep}, and CC \cite{li2021contrastive}. DiEC optimizes $\mathcal{L}_{\mathrm{KL}}+\mathcal{L}_{\mathrm{Gr}}+\mathcal{L}_{\mathrm{En}}$ with the selected COL + COT, and the standard denoising loss $\mathcal{L}_{\mathrm{Re}}$ at random timesteps. Ground-truth labels are reserved solely for evaluation, and the number of clusters $K$ is predefined. Images are normalized for diffusion input.

\paragraph{Main Results.}
Table~\ref{tab:main_acc_nmi_ari} shows that DiEC achieves excellent clustering performance, outperforming embedding-based baselines, including DEC, IDEC, DCEC, VaDE, and ClusterGAN, as well as the diffusion-based ClusterDDPM. Moreover, DiEC outperforms augmentation-consistency methods, including IIC, DCCM, and CC, especially on grayscale benchmarks such as MNIST, USPS, and Fashion-MNIST.

\paragraph{Ablation Study.}
Table~\ref{tab:ablation_super} reports the contribution of each component on the USPS dataset. First, compared with randomly selection of a layer and timestep (ID 1), the identified COL (ID 2) provides a 4.6\% improvement in ACC. Next, by fixing the timestep to COT, ID 3 brings a large improvement in clustering performance, resulting in an ACC of 96.83\%. This confirms the importance of layer and timestep selection, and the effectiveness of the smoothed Scott Score serving as a reliable evaluation metric for locating COL and COT. Then, the subsequent integration of KL self-training, residual decoupling, and adaptive graph regularization (ID 4-6) further refines the feature space, thereby enhancing the clustering performance. Finally, the employment of random-timestep diffusion consistency (ID 7) further refines the representation, culminating in the performance of 98.49\% in ACC, 96.95\% in NMI, and 95.65\% in ARI.
\setcounter{table}{2}
\begin{table}[H]
    \centering
    \small
    \renewcommand{\arraystretch}{1.25}
    \setlength{\tabcolsep}{3pt}

\caption{\textbf{Ablation analysis on USPS.} 
Starting from a random baseline averaged over 50 runs (random layer and timestep), we incrementally add COL/COT selection and the DiEC objectives. The COL variant is also averaged over 50 runs by fixing the layer to COL while randomly sampling timesteps. Here, \textbf{Res.} denotes Residual Decoupling. The full model (ID~7) achieves superior performance.}
    \label{tab:ablation_super}

    \resizebox{\columnwidth}{!}{%
    \begin{tabular}{
        c l
        S[table-format=2.2] @{\hspace{0.35em}} >{\raggedleft\arraybackslash}p{2.6em}
        S[table-format=2.2] @{\hspace{0.35em}} >{\raggedleft\arraybackslash}p{2.6em}
        S[table-format=2.2] @{\hspace{0.35em}} >{\raggedleft\arraybackslash}p{2.6em}
    }
        \toprule
        \rowcolor{rowgray}
        \textbf{ID} & \textbf{Methods}
        & \multicolumn{2}{c}{\textbf{ACC(\%)}}
        & \multicolumn{2}{c}{\textbf{NMI(\%)}}
        & \multicolumn{2}{c}{\textbf{ARI(\%)}}
        \\
        \cmidrule(lr){3-4}\cmidrule(lr){5-6}\cmidrule(lr){7-8}
        \rowcolor{white}
        & & \textbf{Val.} & \textbf{$\Delta$}
          & \textbf{Val.} & \textbf{$\Delta$}
          & \textbf{Val.} & \textbf{$\Delta$}
        \\
        \midrule
        \rowcolor{rowgray}
        1 & Baseline: Random $t$ and $l$
          & 40.64 & {\textcolor{gray}{--}}
          & 25.67 & {\textcolor{gray}{--}}
          & 29.60 & {\textcolor{gray}{--}} \\

        \rowcolor{white}
        2 & \textbf{+} Selected $l^*$(COL)
          & 45.21 & \up{4.57}
          & 28.04 & \up{2.37}
          & 30.39 & \up{0.79} \\

        \rowcolor{rowgray}
        3 & \textbf{+} Selected $t^*$ (COT)
          & 96.83 & \up{51.62}
          & 93.82 & \up{65.78}
          & 92.21 & \up{61.82} \\

        \rowcolor{white}
        4 & \textbf{+} Direct Fine-tuning ($\mathcal{L}_{\text{KL}}$)
          & 97.11 & \up{0.28}
          & 94.31 & \up{0.49}
          & 92.66 & \up{0.45} \\

        \rowcolor{rowgray}
        5 & \textbf{+} Residual Decoupling (Res.)
          & 97.21 & \up{0.10}
          & 94.46 & \up{0.15}
          & 92.83 & \up{0.17} \\

        \rowcolor{white}
        6 & \textbf{+} Graph Regularization ($\mathcal{L}_{\text{Gr}} + \mathcal{L}_{\text{En}}$)
          & 98.37 & \up{1.16}
          & 96.76 & \up{2.30}
          & 95.45 & \up{2.62} \\

        \rowcolor{rowgray}
        7 & \textbf{+} Diffusion Consistency ($\mathcal{L}_{\text{Re}}$) \;\;
          & {\bfseries 98.49} & \up{0.12}
          & {\bfseries 96.95} & \up{0.19}
          & {\bfseries 95.65} & \up{0.20} \\

        \midrule
        \rowcolor{rowgray}
        \multicolumn{2}{l}{\textbf{$\Delta$ (Full $-$ Baseline)}} &
        \multicolumn{1}{c}{\textcolor{gray}{--}} & \up{57.85} &
        \multicolumn{1}{c}{\textcolor{gray}{--}} & \up{71.28} &
        \multicolumn{1}{c}{\textcolor{gray}{--}} & \up{66.05} \\
        \bottomrule

    \end{tabular}%
    }
\end{table}

\paragraph{Qualitative Analysis.}
Fig.~\ref{fig:usps_tsne_1x4} visualizes the evolution of USPS embeddings via t-SNE, utilizing features extracted from the COL. Fig.~\ref{fig:usps_tsne_1x4} (a) and Fig.~\ref{fig:usps_tsne_1x4} (b) display the results at diffusion timestep $t{=}1$. Despite the convergence of pretraining after 400 epochs, the cluster boundaries remain relatively ambiguous. In contrast, switching to the identified COT leads to more dispersed class distributions Fig.~\ref{fig:usps_tsne_1x4} (c). Finally, the full DiEC fine-tuning at the COT further refines the structure, yielding highly separable clusters Fig.~\ref{fig:usps_tsne_1x4} (d).

\section{Conclusion}
In this paper, DiEC, a diffusion-based deep clustering framework, is proposed to directly leverage clustering-friendly embedding representations from pretrained diffusion models for unsupervised clustering. A complementary optimal search strategy is therefore designed to recognize COL and COT in the layer$\times$timestep space through an unsupervised Scott Score criterion. Then, DiEC performs clustering on extracted representations with DEC-style KL-divergence objective, supplemented by a random-timestep diffusion denoising objective to maintain the generative capability of pretrained diffusion models. Experiments and ablation studies on four benchmarks demonstrate the superiority of DiEC over embedding-based and augmentation-consistency clustering methods, revealing that the effective feature selection based on identified COL + COT is crucial for improving clustering performance.

\section*{Acknowledgments}
This work was supported in part by the National Natural Science Foundation of China under Grants No. 62273164 and No. 62373164, the Project of Central Government Guides Local Program under Grant  YDZX2024075, the Taishan Scholars Program of Shandong Province under Grant tsqn202507271, and the Taishan Experts Program under Grant tscy20241154.
\clearpage

\appendix
\setcounter{figure}{0}
\setcounter{algorithm}{0}
\renewcommand{\thefigure}{\Alph{section}.\arabic{figure}}
\renewcommand{\thealgorithm}{\Alph{section}.\arabic{algorithm}}
\renewcommand{\thefigure}{A\arabic{figure}}
\renewcommand{\thetable}{A\arabic{table}}
\renewcommand{\thealgorithm}{A\arabic{algorithm}}
\renewcommand{\theequation}{A\arabic{equation}}
\section{Optimal Search algorithm}
\label{app:A}
Optimal Search adopts a two-stage strategy. In both stages, we evaluate candidate representations on a sampled subset and reduce diffusion stochasticity by averaging over multiple noise trials, as shown in Algorithm~\ref{alg:os}.

In Stage~1, we sample a subset $\mathcal{D}_s$ and evaluate each layer over a timestep set $\mathcal{T}$. We then smooth the scores across timesteps and compute a layer-level score by averaging the top-$\rho$ fraction of each layer's timestep-level scores, which is used to select the $\mathrm{COL}$.

In Stage~2, we fix the selected $\mathrm{COL}$ and evaluate the same timestep set $\mathcal{T}$ to identify the $\mathrm{COT}$, again using smoothed scores for robust selection, and we stop early if the score does not improve for $P$ consecutive timesteps.

\begin{figure*}[!t]
  \centering
  \begin{minipage}[t]{0.49\linewidth}
    \centering
    \includegraphics[width=0.75\linewidth]{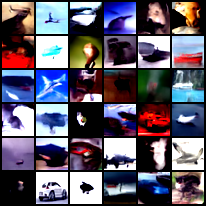}
    \vspace{-1.2mm}
    {\footnotesize\textbf{\\(a) Pretrained diffusion model}}
  \end{minipage}\hfill
  \begin{minipage}[t]{0.49\linewidth}
    \centering
    \includegraphics[width=0.75\linewidth]{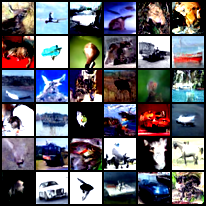}
    \vspace{-1.2mm}
    {\footnotesize\textbf{\\(b) After DiEC fine-tuning}}
  \end{minipage}

  \vspace{-1.5mm}
  \caption{\textbf{Qualitative samples on CIFAR-10 (before vs.\ after DiEC).}
  The two grids are generated using the same sampling configuration, and the diffusion model’s generative capability is well preserved after fine-tuning, supporting the role of $\mathcal{L}_{\mathrm{Re}}$ in maintaining generative consistency.}
  \label{fig:supp_cifar10_samples_before_after}
  \vspace{-2mm}
\end{figure*}
\begin{algorithm}[H]
\caption{Optimal Search for COL and COT}
\label{alg:os}
\textbf{Input}: Dataset $\mathcal{D}$; pretrained feature extractor $\{h_{\theta}^{l}\}_{l\in\mathcal{L}}$;
timestep range $[1,T_s]$; stride $\Delta t$; subset size $m$; noise trials $R$;
PCA dim $d$; smoothing window $w$; top fraction $\rho$; patience $P$
\\
\textbf{Output}: COL $l^{*}$, COT $t^{*}$
\begin{algorithmic}[1]
\STATE Sample subset $\mathcal{D}_s \subset \mathcal{D}$ with $|\mathcal{D}_s|=m$
\STATE $\mathcal{T} \leftarrow \{1,\,1+\Delta t,\,1+2\Delta t,\,\dots,\,T_s\}$

\vspace{0.5mm}
\STATE \textbf{Stage 1: layer selection}
\FOR{$l \in \mathcal{L}$}
  \FOR{$t \in \mathcal{T}$}
    \FOR{each $x_{0,i}\in\mathcal{D}_s$}
      \STATE Sample noises $\{\epsilon^{(r)}\}_{r=1}^{R}$ and form $\{x_{t,i}^{(r)}\}_{r=1}^{R}$
      \STATE $e_{i}^{\, l,(r)}(t)\leftarrow h_{\theta}^{l}\!\left(x_{t,i}^{(r)},t\right)$
      \STATE $\bar{e}_{i}^{\, l}(t)\leftarrow \frac{1}{R}\sum_{r=1}^{R} e_{i}^{\, l,(r)}(t)$
    \ENDFOR
    \STATE $\mathbf{E}_{l,t}\leftarrow \{\bar{e}_i^{\, l}(t)\}_{x_{0,i}\in\mathcal{D}_s}$
    \STATE $\tilde{\mathbf{E}}_{l,t}\leftarrow \ell_2\text{-Norm}\!\big(\mathrm{PCA}_d(\mathbf{E}_{l,t})\big)$
    \STATE $\widetilde{\mathrm{SS}}(l,t)\leftarrow \mathrm{Scott}\!\big(\tilde{\mathbf{E}}_{l,t}\big)$
  \ENDFOR
  \STATE $\widetilde{\mathrm{SS}}_{Sm}(l,t)\leftarrow \mathrm{MovAvg}\!\big(\widetilde{\mathrm{SS}}(l,t), w\big)$
  \STATE $\widehat{S}_{\rho}(l)\leftarrow \mathrm{MeanTop}_{\rho}\!\big(\widetilde{\mathrm{SS}}_{Sm}(l,t)\big)$
\ENDFOR
\STATE $l^{*}\leftarrow \arg\max_{l\in\mathcal{L}} \widehat{S}_{\rho}(l)$

\vspace{0.5mm}
\STATE \textbf{Stage 2: timestep selection}
\STATE $S_{\max} \leftarrow -\infty$; \ $t^{*} \leftarrow 0$; \ $c \leftarrow 0$ 
\FOR{$t \in \mathcal{T}$}
  \FOR{each $x_{0,i}\in\mathcal{D}_s$}
    \STATE Sample noises $\{\epsilon^{(r)}\}_{r=1}^{R}$ and form $\{x_{t,i}^{(r)}\}_{r=1}^{R}$
    \STATE $e_{i}^{(r)}(t)\leftarrow h_{\theta}^{l^{*}}\!\left(x_{t,i}^{(r)},t\right)$
    \STATE $\bar e_{i}(t)\leftarrow \frac{1}{R}\sum_{r=1}^{R} e_{i}^{(r)}(t)$
  \ENDFOR
  \STATE $\mathbf{E}_{{l^*,t}}\leftarrow \{\bar e_i(t)\}_{x_{0,i}\in\mathcal{D}_s}$
  
  \vspace{0.5mm}
  \STATE ${SS}(l^{*},t)\leftarrow \mathrm{Scott}\!\big(\mathbf{E}_{l^*,t}\big)$
  \STATE ${SS}_{Sm}(l^{*},t)\leftarrow \mathrm{OnlineMovAvg}({SS}(l^{*},t), t, w)$
  \IF{${SS}_{Sm}(l^{*},t) > S_{\max}$}
    \STATE $S_{\max} \leftarrow {SS}_{Sm}(l^{*},t)$; \ $t^{*} \leftarrow t$
    \STATE $c \leftarrow 0$
  \ELSE
    \STATE $c \leftarrow c + 1$
  \ENDIF
  
\IF{$c \ge P$}
    \STATE \textbf{break}
\ENDIF
\ENDFOR
\RETURN $(l^{*},t^{*})$
\end{algorithmic}
\end{algorithm}

\newpage
\section{Qualitative Comparison of Generation}
\label{app:B}
We evaluated the generative capability of the proposed DiEC on CIFAR-10. Fig.~\ref{fig:supp_cifar10_samples_before_after} presents images generated by the pretrained and fine-tuned models from the same set of samples. Both models exhibit high image fidelity and diversity. Notably, the images generated by the fine-tuned DiEC contain more details, which validates the effectiveness of our reconstruction objective $\mathcal{L}_{\text{Re}}$  in maintaining generative consistency.
\clearpage
\bibliographystyle{named}
\bibliography{ijcai26}
\end{document}